\documentclass[11pt,a4paper]{article}
\usepackage[hyperref]{acl2021}
\usepackage{times}
\usepackage{latexsym}

\usepackage{microtype}

\aclfinalcopy % Uncomment this line for the final submission
\def\aclpaperid{902} %  Enter the acl Paper ID here

\usepackage{multirow}
\usepackage{multicol}
\usepackage{amsmath} % used for boldsymbol.
\renewcommand{\vec}[1]{\boldsymbol{#1}} % Uncomment for BOLD vectors.
\usepackage{amssymb}% http://ctan.org/pkg/amssymb
\usepackage{pifont}% http://ctan.org/pkg/pifont
\usepackage{color, soul} 
\usepackage[switch]{lineno}  %
\usepackage{graphicx}

\title{Capturing Event Argument Interaction via A Bi-Directional Entity-Level Recurrent Decoder}

\author{Xi Xiangyu$^{1,2}$, Wei Ye$^{1,}$\footnotemark[2] , Shikun Zhang$^{1,}$\footnotemark[2] , Quanxiu Wang$^{3}$, Huixing Jiang$^{2}$, Wei Wu$^{2}$ \\
  $^1$ National Engineering Research Center for Software Engineering, Peking University, \\Beijing, China \\
  $^2$ Meituan Group, Beijing, China \\
  $^3$ RICH AI, Beijing, China \\
  \texttt{\{xixy,wye,zhangsk\}@pku.edu.cn} \\}

\date{}

\begin{document}
\maketitle
\renewcommand{\thefootnote}{\fnsymbol{footnote}}
\footnotetext[2]{Corresponding authors.}
\begin{abstract}

Capturing interactions among event arguments is an essential step towards robust event argument extraction (EAE). However, existing efforts in this direction suffer from two limitations: 1) The argument role type information of contextual entities is mainly utilized as training signals, ignoring the potential merits of directly adopting it as semantically rich input features; 2) The argument-level sequential semantics, which implies the overall distribution pattern of argument roles over an event mention, is not well characterized. To tackle the above two bottlenecks, we formalize EAE as a Seq2Seq-like learning problem for the first time, where a sentence with a specific event trigger is mapped to a sequence of event argument roles. A neural architecture with a novel Bi-directional Entity-level Recurrent Decoder (BERD) is proposed to generate argument roles by incorporating contextual entities' argument role predictions, like a word-by-word text generation process, thereby distinguishing implicit argument distribution patterns within an event more accurately.

\end{abstract}

\renewcommand{\thefootnote}{\arabic{footnote}}
\section{Introduction}
\label{Introduction}
% 介绍事件抽取
% Event extraction (EE), which aims to extract events with specific types and their participants and attributes from unstructured data  \cite{nguyen:2015event}, is an important and challenging task in information extraction. 
% Event extraction is an important and challenging task of information extraction in natural language processing.
Event argument extraction (EAE), which aims to identify the entities serving as event arguments and classify the roles they play in an event, is a key step towards event extraction (EE).
For example, given that the word ``fired'' triggers an \emph{Attack} event in the sentence ``In Baghdad, a cameraman died when an American tank fired on the Palestine Hotel''
, EAE need to identify that ``Baghdad'', ``cameraman'', ``American tank'', and ``Palestine hotel'' are arguments with \emph{Place}, \emph{Target}, \emph{Instrument}, and \emph{Target} as roles respectively.

Recently, deep learning models have been widely applied to event argument extraction and achieved significant progress\cite{chen:2015event,nguyen:2016joint,sha:2018jointly,yang2019exploring,wang2019hmeae,zhang2020two,du2020event}.
% including convolutional neural network (CNN)  \cite{chen:2015event,chen:2017freebase} and recurrent neural network (RNN)  \cite{nguyen:2016joint,zeng:2016convolution,sha:2018jointly}, 
Many efforts have been devoted to improving EAE by better characterizing argument interaction, categorized into two paradigms. The first one, named ~\textbf{inter-event argument interaction} in this paper, concentrates on mining information of the target entity (candidate argument) in the context of other event instances ~\cite{hong:2011using,nguyen:2016joint}, e.g., the evidence that a \emph{Victim} argument for the \emph{Die} event is often the \emph{Target} argument for the \emph{Attack} event in the same sentence. The second one is ~\textbf{intra-event argument interaction}, which exploits the relationship of the target entity with others in the same event instance~\cite{hong:2011using,sha2016rbpb,sha:2018jointly}.
We focus on the second paradigm in this paper.

Despite their promising results, existing methods on capturing intra-event argument interaction suffer from two bottlenecks. 

~\textbf{(1) The argument role type information of contextual entities is underutilized.}  As two representative explorations, dBRNN \cite{sha:2018jointly} uses an intermediate tensor layer to capture latent interaction between candidate arguments; RBPB~\cite{sha2016rbpb}  estimates whether two candidate argument belongs to one event or not, serving as constraints on a Beam-Search-based prediction algorithm. Generally, these works use the argument role type information of contextual entities as ~\textbf{auxiliary supervision signals for training} to refine input representation. However, one intuitive observation is that the argument role types can be utilized straightforwardly as semantically rich ~\textbf{input features}, like how we use entity type information. To verify this intuition, we conduct an experiment on ACE 2005 English corpus, in which CNN  ~\cite{nguyen2015event} is utilized as a baseline. For an entity, we incorporate the ground-truth roles of its contextual arguments into the baseline model's input representation, obtaining model CNN(w. role type). As expected, CNN(w. role type) outperforms CNN significantly as shown in Table \ref{tab:argument_pre_results}\footnote{In the experiment we skip the event detection phase and directly assume all the triggers are correctly recognized.}.

\begin{table}[hbt]
	\centering
	\begin{tabular}{lccc}
		\hline 
		Model &$P$ & $R$ & $F_1$\\
		\hline
		CNN & 57.8 &	55.0 &	56.3 \\
% 		\hline
% 		CNN(Left) & 58.4 &	59.1 &	58.7 \\
% 		\hline
		CNN(w. role type) & 59.8 & 60.3 &	60.0 \\
		\hline
	\end{tabular}
	\caption{Experimental results of CNN and its variant on ACE 2005.}
	\label{tab:argument_pre_results}
\end{table}

The challenge of the method lies in knowing the ground-truth roles of contextual entities in the inference (or testing) phase. That is one possible reason why existing works do not investigate in this direction.  Here we can simply use predicted argument roles to approximate corresponding ground truth for inference. We believe that the noise brought by prediction is tolerable, considering the stimulating effect of using argument roles directly as input.

~\textbf{(2) The distribution pattern of multiple argument roles within an event is not well characterized.}  For events with many entities, distinguishing the overall appearance patterns of argument roles is essential to make accurate role predictions. In dBRNN \cite{sha:2018jointly}, however, there is no specific design involving constraints or interaction among multiple prediction results, though the argument representation fed into the final classifier is enriched with synthesized information (the tensor layer) from other arguments. RBPB \cite{sha2016rbpb} explicitly leverages simple correlations inside each argument pair, ignoring more complex interactions in the whole argument sequence. Therefore, we need a more reliable way to learn the sequential semantics of argument roles in an event.

% To explicitly utilize the argument role information of other entities, one straightforward way is to feed the prediction result of the previously processed entity to the processing unit of the current entity recurrently. 

To address the above two challenges, we formalize EAE as a  Seq2Seq-like learning problem  \cite{bahdanau2014neural} of mapping a sentence with a specific event trigger to a sequence of event argument roles.  To fully utilize both left- and right-side argument role information, inspired by the bi-directional decoder for machine translation  \cite{zhang2018asynchronous},  we propose a neural architecture with a novel Bi-directional Entity-level Recurrent Decoder (BERD) to generate event argument roles entity by entity. The predicted argument role of an entity is fed into the decoding module for the next or previous entity recurrently like a text generation process. In this way, BERD can identify candidate arguments in a way that is more consistent with the implicit distribution pattern of multiple argument roles within a sentence, similar to text generation models that learn to generate word sequences following certain grammatical rules or text styles.

% 介绍实验

% We have conducted extensive experiment on the ACE 2005 dataset and the experimental results demonstrate that BERD 
% can fuse argument information from left- and right-side effectively and outperforms state-of-the-art baselines. 
% 本文的贡献总结
The contributions of this paper are:

% The superiority of BERD is more significant when there are more entities in a sentence.

\begin{enumerate}
    \item We formalize the task of event argument extraction as a Seq2Seq-like learning problem for the first time, where a sentence with a specific event trigger is mapped to a sequence of event argument roles.
    \item We propose a novel architecture with a Bi-directional Entity-level Recurrent Decoder (BERD) that is capable of leveraging the argument role predictions of left- and right-side contextual entities and distinguishing argument roles' overall distribution pattern.
    \item Extensive experimental results show that our proposed method outperforms several competitive baselines on the widely-used ACE 2005 dataset. 
    % clearly set up a new state-of-the-art result on the widely-used ACE 2005 dataset. 
    BERD's superiority is more significant given more entities in a sentence.
\end{enumerate}

\section{Problem Formulation}

Most previous works formalize EAE as either a word-level sequence labeling problem \cite{nguyen:2016joint,zeng:2016convolution,yang2019exploring} or an  entity-oriented classic classification problem \cite{chen:2015event,wang2019hmeae}.
We formalize EAE as a Seq2Seq-like learning problem as follows.
Let $S=\{w_{1},...,w_{n}\}$ be a sentence where $n$ is the sentence length and $w_i$ is the $i$-th token. Also, let $E=\{e_1,...,e_k\}$ be the entity mentions in the sentence where $k$ is number of entities. 
Given that an event triggered by $t \in S$ is detected in ED stage
% and each entity $e_i$ in the sentence is regarded as candidate argument
, EAE need to map the sentence with the event to a sequence of argument roles $R=\{y_1,...,y_k\}$, where $y_i$ denotes the argument role that entity $e_i$ plays in the event.

% Given that $m$ events are detected in ED stage and denoted by $T=\{t_1,t_2,...,t_m\}$, EAE need to predict the role $r_{i,j}$ that each entity $e_i$ plays in each event $t_j$, where $r_{i,j} \in R$.

\begin{figure*}[hbt]
    \centering
    \includegraphics[scale=0.46]{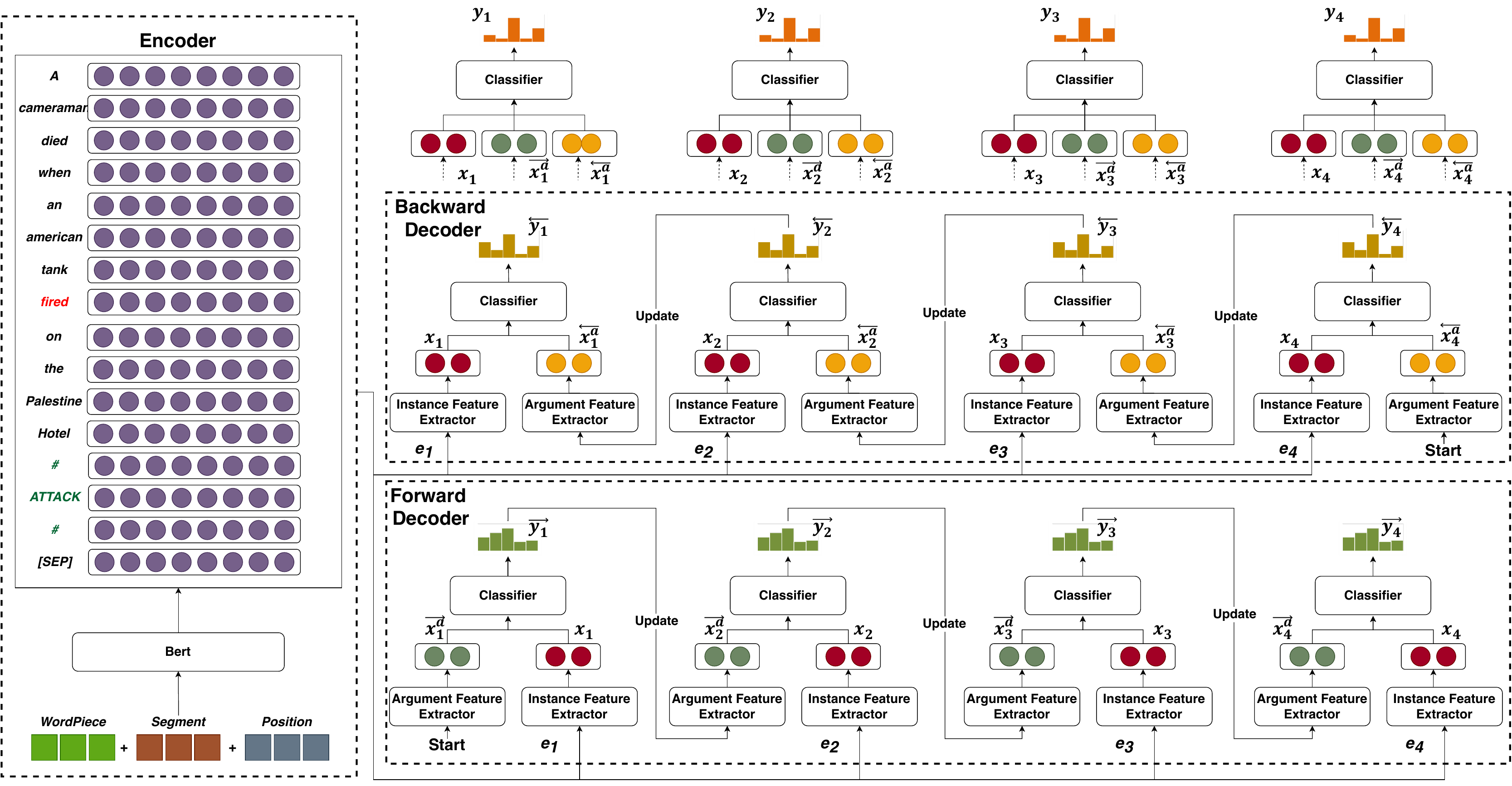}
    \caption{The detailed architecture of our proposed approach. The figure depicts a concrete case where a sentence contains an \emph{Attack} event (triggered by ``fired'') and 4 candidate arguments $\{e_1,e_2,e_3,e_4\}$. The encoder on the left converts the sentence into intermediate continuous representations. Then the forward decoder and backward decoder generates the argument roles sequences in a left-to-right and right-to-left manner (denoted by $\overrightarrow{\vec{y}_i}$ and $\overleftarrow{\vec{y}_i}$) respectively. A classifier is finally adopted to make the final prediction $\vec{y}_i$.
    The forward and backward decoder shares the instance feature extractor and generate the same instance representation $\vec{x_i}$ for $i$-th entity.
    The histogram in green and brown denotes the probability distribution generated by forward decoder and backward decoder respectively. The orange histogram denotes the final predictions. Note that the histograms are for illustration only and do not represent the true probability distribution.
    }
    \label{fig:model}
\end{figure*}

\section{The Proposed Approach}

We employ an encoder-decoder architecture for the problem defined above, which is similar to most Seq2Seq models in machine translation \cite{vaswani2017attention,zhang2018asynchronous}, automatic text summarization \cite{song2019abstractive,shi2021neural}, and speech recognition \cite{tuske2019advancing,hannun2019sequence} from a high-level perspective. 

In particular, as Figure \ref{fig:model} shows, our architecture consists of an encoder that converts the sentence $S$ with a specific event trigger into intermediate vectorized representation and a decoder that generates a sequence of argument roles entity by entity. The decoder is an entity-level recurrent network whose number of decoding steps is fixed, the same as the entity number in the corresponding sentence. On each decoding step, we feed the prediction results of the previously-processed entity into the recurrent unit to make prediction for the current entity. Since the predicted results of both left- and right-side entities can be potentially valuable information,we further incorporate a bidirectional decoding mechanism that integrates a forward decoding process and a backward decoding process effectively. 

\subsection{Encoder}
Given the sentence $S=$ ($w_{1},...,w_{n}$) containing a trigger $t \in S$ and $k$ candidate arguments $E=\{e_1,...,e_k\}$, an encoder is adopted to encode the word sequence into a sequence of continuous representations as follows\footnote{The trigger word can be detected by any event detection model, which is not the scope of this paper.},
\begin{equation}
    \vec{H}=(\vec{h}_1,...,\vec{h}_n) = F(w_{1},...,w_{n})
\end{equation}
where $F(\cdot)$ is the neural network to encode the sentence. In this paper, we select BERT \cite{devlin2019bert} as the encoder. Considering representation $\vec{H}$ does not contain event type information, which is essential for predicting argument roles. 
% There is no straightforward way to incorporate event type information into BERT's input representation of a word , which is constructed by summing embeddings of the corresponding token, segment, and position. 
We append a special phrase denoting event type of $t$ into each input sequence, such as ``\# ATTACK \#''.
% shown in Figure \ref{fig:model}.
% as the following example shows:\\
% [ClS] In Baghdad, a cameraman died when an American tank fired on the Palestine Hotel \textbf{\# ATTACK \# }[SEP]
\subsection{Decoder}
Different from traditional token-level Seq2Seq models, we use a bi-directional entity-level recurrent decoder (BERD) with a classifier to generate a sequence of argument roles entity by entity. BERD consists of a forward and backward recurrent decoder, which exploit the same recurrent unit architecture as follows. 

% In decoding stage, we first use a Bi-directional entity-level recurrent decoder which consists of a forward and backward recurrent decoder generating argument roles sequence in a left-to-right and right-to-left manner respectively.
% We propose a Bi-directional entity-level recurrent decoder with a classifier in the decoding stage.
% In decoding stage, we use a Bi-directional entity-level recurrent decoder with a classifier
% The decoder 
% We use a bi-directional entity-level recurrent decoder with a classifier in the decoding stage.

% The decoder consists of a forward recurrent decoder and a backward recurrent decoder. The two decoders exploit the same recurrent unit, which is used to generate the argument role for an entity.
% Given the instance above, the decoder aims to predict what role argument candidate $a$ plays in the event triggered by $t$. To explicitly utilize the argument role information of other entities, we design an entity-level recurrent decoder which is composed of recurrent units enabling exploitation of such information. 
\subsubsection{Recurrent Unit}
% The recurrent unit is explicitly designed to simultaneously utilize two kind of information: (1) the instance information which contains the corresponding sentence, event and argument candidate (denoted by $S,t,e$); (2) contextual argument information which consists of argument roles of other entities (denoted by $A$). To be specific, $A$ is a list recording argument roles for each token in $S$. Previously-predicted entities are represented by generated labels. The candidate entity $e$ is labeled as a special label ``To-Predict''. Other entities and tokens that don't collapse with entities are represented as ``N/A''.
% The recurrent unit exploits two feature extractors to extract features as follows:

The recurrent unit is designed to explicitly utilize two kinds of information: (1) the instance information which contains the sentence, event, and candidate argument (denoted by $S,t,e$); and (2) contextual argument information which consists of argument roles of other entities (denoted by $A$). The recurrent unit exploits two corresponding feature extractors as follows:\\
\textbf{Instance Feature Extractor.} 
Given the representation $\vec{H}$ generated by encoder, dynamic multi-pooling  \cite{chen:2015event} is then applied to extract max values of three split parts, which are decided by the event trigger and the candidate argument. The three hidden embeddings are aggregated into an instance feature representation $\vec{x}$ as follows:
\begin{equation}
\begin{split}
    [\vec{x}_{1,p_t}]_i &= \max \{[\vec{h}_1]_i,...,[\vec{h}_{pt}]_i\} \\
    [\vec{x}_{p_t+1,p_e}]_i &= \max \{[\vec{h}_{p_t+1}]_i,...,[\vec{h}_{pe}]_i\} \\
    [\vec{x}_{p_e+1,n}]_i &= \max \{[\vec{h}_{p_e+1}]_i,...,[\vec{h}_{n}]_i\} \\
    \vec{x} & = [\vec{x}_{1,p_t}; \vec{x}_{p_t+1,p_e}; \vec{x}_{p_e+1,n}]
\end{split}
\label{equ:piecewisepooling}
\end{equation}
where $[\cdot]_i$ is the $i$-th value of a vector, $p_t$, $p_e$ are the positions of trigger $t$ and candidate argument $e$
\footnote{Equation \ref{equ:piecewisepooling} assumes that the entity mention lies after the trigger. If the entity mention lies before the trigger, we switch $p_t$ and $p_e$ in the equation to get a right split.}. \\
% We concatenate the piece-wise max-pooling results as the instance feature representation $\vec{x}$.
\textbf{Argument Feature Extractor.} To incorporate previously-generated arguments, we exploit CNN network to encode the instance with arguments information as follows.
% In this paper, we exploit CNN network to extract argument features as follows.

\textbf{\emph{Input.}} Different from \citet{chen:2015event} where input embedding of each word consists of its word embedding, position embedding, and event type embedding, we append the embedding of argument roles into the input embedding for each word by looking up the vector $A$, which records argument role for each token in $S$.
% A, which is a sequence of argument roles for each token in $S$. 
% In this paper, we use vector $A$ to denote argument roles for each token in $S$
% according to $A$.
% To be specific, $A$ is a sequence of argument roles for each token in $S$. 
In $A$, tokens of previously-predicted arguments are assigned with the generated labels, tokens of the candidate entity $e$ are assigned with a special label ``To-Predict'', and the other tokens are assigned with label \emph{N/A}. 

\textbf{\emph{Convolution.}}The convolution layer is applied to encode the word sequence into hidden embeddings:
\begin{equation}
    (\vec{h}^{a}_1,,...,\vec{h}^a_n) = {\rm CNN} (w_{1},...,t,...,e,...,w_{n})
\end{equation}
where the upperscript $a$ denotes argument.

\textbf{\emph{Pooling.}} Max-pooing operation is then applied to extract the argument feature $\vec{x}^a$ as follows,
\begin{equation}
    \vec{x}^a = {\rm MaxPooling}(\vec{h}^{a}_1,...,\vec{h}^a_n)
\end{equation}

We concatenate the instance feature representation $\vec{x}$ and the argument feature representation $\vec{x}^a$ as the input feature representation for the argument role classifier, and estimate the role that $e$ plays in the event as follows:
% the probability of instance $e$ being of $i$-th argument role as follows:
\begin{equation}
\begin{split}
    \vec{p} & = f(\vec{W}[\vec{x};\vec{x}^a] + \vec{b}) \\
    \vec{o} & = {\rm Softmax}(\vec{p})
    % p(i|e;Env,S,t) &= \frac{e^{\vec{o}_i}}{\sum_{j=1}^{n_r}{\vec{o}_j}}
    % p(y^i|x,\theta) & = \frac{e^{\vec{o}_i}}{\sum_{j=1}^{n_r}{\vec{o}_j}}
\end{split}    
\end{equation}
where $\vec{W}$ and $\vec{b}$ are weight parameters. $\vec{o}$ is the probability distribution over the role label space.

% Each dimension $\vec{p}_i$ of the vector $\vec{p}$ represents the probability of $e$ playing argument role $r_i$ in the event triggered by $t$. 
For the sake of simplicity, in rest of the paper we use ${\rm Unit}(S,t,e,A)$ to represent the calculation of probability distribution $\vec{o}$ by recurrent unit with $S,t,e,A$ as inputs.

\subsubsection{Forward Decoder}
Given the sentence $S$ with $k$ candidate arguments $E=\{e_1,...,e_k\}$,
the forward decoder exploits above recurrent unit and generates the argument roles sequence in a left-to-right manner.
% Given the instance embedding $\vec{x}$ of the instance encoder, 
The conditional probability of the argument roles sequence is calculated as follows:
\begin{equation}
    P(R|E,S,t) = \prod_{i=1}^{k}{p(y_i|e_i;R_{<i},S,t)}
\end{equation}
where $R_{<i}$ denotes the role sequence $\{y_{1},...,y_{i-1}\}$ for $\{e_1,...,e_{i-1}\}$.

For $i$-th entity $e_i$, the recurrent unit generates prediction as follows:
\begin{equation}
    % \overleftarrow{y_i} = 
    % p(y_i|e_i;R_{<i},S,t) 
    \overrightarrow{\vec{y}_i}= {\rm Unit}(S,t,e_i,\overrightarrow{A_i})
\end{equation}
where $\overrightarrow{\vec{y}_i}$ denotes the probability distribution over label space for $e_i$ and $\overrightarrow{A_i}$ denotes the contextual argument information of $i$-th decoding step, which contains previously-predicted argument roles $R_{<i}$.
% \begin{equation}
% \begin{split}
%     \vec{o} & = W[\vec{x};\vec{x}^e] \\
%     p(r_i|e_i;R_{<i},W,t) &= \frac{e^{\vec{o}_i}}{\sum_{j=1}^{n_r}{\vec{o}_j}}
%     % p(y^i|x,\theta) & = \frac{e^{\vec{o}_i}}{\sum_{j=1}^{n_r}{\vec{o}_j}}
% \end{split}
% \end{equation}
Then we update $\overrightarrow{A_{i+1}}$ by labeling $e_i$ as $g(\overrightarrow{\vec{y}_i})$ for next step $i$+1,
% \begin{equation}
%     \overrightarrow{A_{i+1}}=\overrightarrow{A_{i}} \cup \{e_i:g(\overrightarrow{\vec{y}_i})\}
% \end{equation}
where $g(\overrightarrow{\vec{y}_i})$ denotes the label has the highest probability under the distribution $\overrightarrow{\vec{y}_i}$.
The argument feature extracted by recurrent units of forward decoder is denoted as $\overrightarrow{\vec{x}}^a_i$. 
% We use $\overrightarrow{p(y_i|e_i)}$ to represent the probability of $e_i$ playing role $y_i$ estimated by forward decoder.

\subsubsection{Backward Decoder}
The backward decoder is similar to the forward decoder, except that it performs decoding in a right-to-left way as follows:
\begin{equation}
    P(R|E,S,t) = \prod_{i=1}^{k}{p(y_i|e_i;R_{>i},S,t)}
\end{equation}
where $R_{>i}$ denotes the role sequence $\{y_{i+1},...,y_{k}\}$ for $\{e_{i+1},...,e_{k}\}$.
The probability distribution over label space for $i$-th entity $e_i$ is calculated as follows:
\begin{equation}
    \overleftarrow{\vec{y}_i}= {\rm Unit}(S,t,e_i,\overleftarrow{A_{i}})
\end{equation}
where $\overleftarrow{A_i}$ denotes the contextual argument information of $i$-th decoding step, which contains previously-predicted argument roles $R_{>i}$.
We update $\overleftarrow{A_{i-1}}$ by labeling $e_i$ as $g(\overleftarrow{\vec{y}_i})$ for next step $i$-1.
% We update the argument information of $e_i$ with the label $g(\overleftarrow{\vec{y}_i})$ for next step $i$-1 as follows:
% \begin{equation}
%     \overleftarrow{A_{i-1}}=\overleftarrow{A_{i}} \cup \{e_i:g(\overleftarrow{\vec{y}_i})\}
% \end{equation}
 The argument feature extracted by recurrent units of backward decoder is denoted as $\overleftarrow{\vec{x}}^a_i$. 
%  We use $\overleftarrow{p(y_i|e_i)}$ to represent the probability of $e_i$ playing role $y_i$ estimated by backward decoder.

% The probability of instance $e_i$ being of role $r_i$ is calculated as follows:
% \begin{equation}
%     p(r_i|e_i;R_{>i},S,t) = {\rm Unit}(e_i,S,t,R_{>i})
% \end{equation}
% After predicting argument candidate $e_i$, the argument information is updated with the predicted role $r_i$ for next step $i$-1. We use $\overleftarrow{p(e_i|r_i)}$ to represent $p(r_i|e_i;R_{>i},S,t)$ for the sake of simplicity.  The argument feature extracted by recurrent units of forward decoder is denoted as $\overleftarrow{\vec{x}}^e$.

\subsubsection{Classifier}
To utilize both left- and right-side argument information, a classifier is then adopted to combine argument features of both decoders and make final prediction for each entity $e_i$ as follows:
\begin{equation}
    \begin{split}
        \vec{p}_i & = f(\vec{W_c}[\vec{x}_i;\overrightarrow{\vec{x}}^a_i;\overleftarrow{\vec{x}}^a_i] + \vec{b_c}) \\
        \vec{y}_i & = {\rm Softmax}(\vec{p}_i)
    \end{split}
    \label{Equ:classifier}
\end{equation}
where $\vec{y_i}$ denotes the final probability distribution for $e_i$. $\vec{W_c}$ and $\vec{b_c}$ are weight parameters. 
\subsection{Training and Optimization}

As seen, the forward decoder and backward decoder in BERD mainly play two important roles. The first one is to yield intermediate argument features for the final classifier, and the second one is to make the initial predictions fed into the argument feature extractor.  Since the initial predictions of the two decoders are crucial to generate accurate argument features,  we need to optimize their own classifier in addition to the final classifier.

We use $\overrightarrow{p(y_i|e_i)}$ and $\overleftarrow{p(y_i|e_i)}$ to represent the probability of $e_i$ playing role $y_i$ estimated by forward and backward decoder respectively.
$p(y_i|e_i)$ denotes the final estimated probability of $e_i$ playing role $y_i$ by Equation \ref{Equ:classifier}.
% $p(y_i|e_i;R_{\ne i},S,t)$ denotes the final estimated probability of $e_i$ playing role $y_i$, where $R_{\ne i}$ denotes the role sequence $\{...,y_{i-1},y_{i+1},...\}$ for $\{...,e_{i-1},e_{i+1},...\}$.
The optimization objective function is defined as follows:

% \begin{equation}
% J(\theta) = -\sum_{S\in D}^{}{\sum_{t\in S}^{}{\sum_{e_i \in E_S}^{}{(\alpha  p(y_i|e_i) + \beta \overrightarrow{p(y_i|e_i)} + \gamma \overleftarrow{p(y_i|e_i)})}}}
% \end{equation}

\begin{equation}
    \begin{split}
    J(\theta) & = -\sum_{S\in D}\sum_{t\in S}\sum_{e_i \in E_S}{\alpha \log p(y_i|e_i;R_{\ne i},S,t)} \\
    & +  \beta \log \overrightarrow{p(y_i|e_i)} + \gamma \log \overleftarrow{p(y_i|e_i)}
    \end{split}
\end{equation}

\noindent where $D$ denotes the training set and $t\in S$ denotes the trigger word detected by previous event detection model in sentence $S$. $E_S$ represents the entity mentions in $S$. $\alpha$, $\beta$ and $\gamma$ are weights for loss of final classifier, forward decoder and backward decoder respectively.
% forward decoder, backward decoder and overall model respectively. 
% Adam optimization method with minibatchs is adopted to minimize $J(\theta)$.

During training, we apply the teacher forcing mechanism where gold arguments information is fed into BERD's recurrent units, enabling paralleled computation and greatly accelerates the training process. 
Once the model is trained, we first use the forward decoder with a greedy search to sequentially generate a sequence of argument roles in a left-to-right manner. Then, the backward decoder performs decoding in the same way but a right-to-left manner. Finally, the classifier combines both left- and right-side argument features and make prediction for each entity as Equation \ref{Equ:classifier} shows.

\section{Experiments}

\subsection{Experimental Setup}
\subsubsection{Dataset}
Following most works on EAE \cite{nguyen:2016joint,sha:2018jointly,yang2019exploring,du2020event}, we evaluate our models on the most widely-used ACE 2005 dataset, which contains 599 documents annotated with 33 event subtypes and 35 argument roles. 
We use the same test set containing 40 newswire documents, a development set containing 30 randomly selected documents and training set with the remaining 529 documents.
% We use the same data split where 40 newswire documents are used for the test set, 30 randomly selected documents are used for the development set, and the remaining 529 documents form the training set.

We notice \citet{wang2019hmeae} used TAC KBP dataset, which we can not access online or acquire from them due to copyright. We believe experimenting with settings consistent with most related works (e.g., 27 out of 37 top papers used only the ACE 2005 dataset in the last four years) should yield convincing empirical results.
% \footnote{We notice \citet{wang2019hmeae} used TAC KBP dataset, which we cannot access online or acquire from them due to copyright. 
% We believe experimenting with settings consistent with most related works (e.g., 27 out of 37 top papers used only the ACL 2005 dataset in the last four years) should yield convincing empirical results.}. 
% An argument is correctly classified if its offsets, role, related trigger type and trigger's offsets exactly match a reference argument. We report precision(P), recall(R) and f1-score(F1) in the experiment.

% \subsubsection{Evaluation Metrics} An argument is correctly classified if its offsets, role, related trigger type and trigger's offsets exactly match a reference argument. We report precision(P), recall(R) and f1-score(F1) in the experiment.

% Different from prior work\footnote{Our investigation shows that prior works used different data splits (althought documents are partitioned into 529/30/40 documents), which is a severe problem in event extraction community.}, we performed a 5-fold cross-validation on the ACE 2005 dataset. We partitioned 599 files into 5 parts. The file names of each fold can be found online. We chose a different fold each time as the testing set and used the remaining four folds as the training set.

\subsubsection{Hyperparameters} We adopt BERT \cite{devlin2019bert} as encoder and the proposed bi-directional entity-level recurrent decoder as decoder for the experiment.
The hyperparameters used in the experiment are listed.\\
\emph{\textbf{BERT.}} The hyperparameters of BERT are the same as the BERT$_{\rm BASE}$ model\footnote{https://github.com/google-research/bert}. We use a dropout probability of 0.1 on all layers.\\
\emph{\textbf{Argument Feature Extractor.}} Dimensions of word embedding, position embedding, event type embedding and argument role embedding for each token are 100, 5, 5, 10 respectively. We utilize 300 convolution kernels with size 3. The glove embedding\cite{pennington2014glove} are utilized for initialization of word embedding\footnote{https://nlp.stanford.edu/projects/glove/}.\\
%  \cite{pennington2014glove}
\emph{\textbf{Training.}} Adam with learning rate of 6e-05, $\beta_1=0.9$, $\beta_2=0.999$, L2 weight decay of 0.01 and learning rate warmup of 0.1 is used for optimization. We set the training epochs and batch size to 40 and 30 respectively. Besides, we exploit a dropout with rate 0.5 on the concatenated feature representations. The loss weights $\alpha$, $\beta$ and $\gamma$ are set to 1.0, 0.5 and 0.5 respectively.

% \subsubsection{Event Detection Method} Following previous work \cite{wang2019hmeae}, we use a pipelined approach for event extraction and implement DMBERT for event detection.
% The P, R and F1 for trigger classification is 74.1\%, 78.0\% and 76.1\% respectively.

% The hyperparameters of DMBERT and DMCNN are shown in Table \ref{tab:hyperparameter}.
% \begin{table}[hbt]
%     \centering
%     \caption{Hyperparameter settings for our proposed models.}
%     \begin{tabular}{c|c}
%     \hline
%     \textbf{Hyperparameters for Decoder} & \textbf{Values} \\ 
%     Word Embedding Dimension & 100 \\
%     Kernel Size & 3 \\
%     Kernel Num & 300 \\
%     Position Embedding Dimension & 5 \\
%     Event Type Embedding Dimension & 5 \\
%     Argument Role Embedding Dimension & 10 \\
%     \hline    
%     \textbf{Hyperparameters for Objective Function} & \textbf{Values} \\ 
%     $\alpha$ & 0.5 \\
%     $\beta$ & 0.5 \\
%     \hline    
%     \textbf{Hyperparameters for Training} & \textbf{Values} \\ 
%     Learning Rate & 6e-05 \\
%     weight decay rate & 0.01 \\
%     Batch Size  &  30 \\
%     Warmup Rate & 0.1 \\
%     $\beta_1$ & 0.9 \\
%     $\beta_2$ & 0.999 \\
%     Dropout Rate & 0.5 \\
%     Training Epoch & 40 \\
%     \hline
%     \end{tabular}
%     \label{tab:hyperparameter}
% \end{table}

% Different from  \cite{chen:2015event} where input embedding of each word consists of word embedding, position embedding and event type embedding, we append argument role embedding into input embedding for each word. 

\subsection{Baselines} 
We compare our method against the following four baselines. The first two are state-of-the-art models that separately predicts argument without considering argument interaction.
% To verify the effectiveness of our method, w
We also implement two variants of DMBERT utilizing the latest inter-event and intra-event argument interaction method, named BERT(Inter) and BERT(Intra) respectively.
% To verify the effectiveness of our method, we also re-implement two argument interaction methods based on DMBERT.
% We re-implement argument interaction method based on  
% and the latter two are DMBERT enhanced by existing argument interaction method.
\begin{enumerate}
    \item \textbf{DMBERT} which adopts BERT as encoder and generate representation for each entity mention based on dynamic multi-pooling operation\cite{wang2019adversarial}. The candidate arguments are predicted separately.
    \item \textbf{HMEAE} which utilizes the concept hierarchy of argument roles and utilizes hierarchical modular attention for event argument extraction \cite{wang2019hmeae}.
    \item \textbf{BERT(Inter)} which enhances DMBERT with inter-event argument interaction adopted by \citet{nguyen:2016joint}. The memory matrices are introduced to store dependencies among event triggers and argument roles.
    \item \textbf{BERT(Intra)} which incorporates intra-event argument interaction adopted by \citet{sha:2018jointly} into DMBERT. The tensor layer and self-matching attention matrix with the same settings are applied in the experiment.
    % incorporate argument-argument interactions and self-matching attention matrix adopted by \cite{sha:2018jointly} into DMBERT.
\end{enumerate}

% Note that representation of the entity mention is mainly determined by last word of the entity mention, 

Following previous work \cite{wang2019hmeae}, we use a pipelined approach for event extraction and implement DMBERT as event detection model.
% The prediction generated by the same event detection model
The same event detection model is used for all the baselines to ensure a fair comparison. 
% The P, R and F1 for trigger classification is 74.1\%, 78.0\% and 76.1\% respectively.

Note that \citet{nguyen:2016joint} uses the last word to represent the entity mention\footnote{\citet{sha:2018jointly} doesn't introduce the details.}, which may lead to insufficient semantic information and inaccurate evaluation considering entity mentions may consist of multiple words and overlap with each other. We sum hidden embedding of all words when collecting lexical features for each entity mention.

\begin{table}[hbt]
	\centering
	\begin{tabular}{lccc}
		% \cline{3-8} &  & P & R & F & P & R & F\\ \textbf{P\%} & \textbf{R\%} & \textbf{F1\%} \\
		\hline 
		\textbf{Model} &  \textbf{P} & \textbf{R} & \textbf{F1} \\
		\hline
		DMBERT & 56.9 & 57.4 & 57.2 \\
		HMEAE & \textbf{62.2} & 56.6 & 59.3 \\
		BERT(Inter) & 58.4 & 57.1 & 57.8\\
		BERT(Intra) & 56.4 & 61.2 & 58.7 \\
		\hline
		BERD & 59.1 & \textbf{61.5} & \textbf{60.3} \\
		\hline
	\end{tabular}
    \caption{Overall performance on ACE 2005 (\%).}
	\label{table:result1}
\end{table}
\subsection{Main Results}
\label{main_results}

The performance of BERD and baselines are shown in Table \ref{table:result1} (statistically significant with $p<0.05$), from which we have several main observations.
% 达到了sota
(1) Compared with the latest best-performed baseline HMEAE, our method BERD achieves an absolute improvement of 1.0 F1, clearly achieving competitive performance.
% clearly setting up a new state-of-the-art performance. 
% DMBERT加上了其他方法都取得了不错的效果
(2) Incorporation of argument interactions brings significant improvements over vanilla DMBERT. For example, BERT(Intra) gains a 1.5 F1 improvement compared with DMBERT, which has the same architecture except for argument interaction.
% Intra的效果好于Inter
(3) Intra-event argument interaction brings more benefit than inter-event interaction (57.8 of BERT(Inter) v.s. 58.7 of BERT(Intra) v.s. 60.3 of BERD).
% BERD
(4) Compared with BERT(Inter) and BERT(Intra), our proposed BERD achieves the most significant improvements.
We attribute the solid enhancement to BERD's novel seq2seq-like architecture that effectively exploits the argument roles of contextual entities.

\subsection{Effect of Entity Numbers}

% 这里需要一个过渡，为什么要分析Entity Number
To further investigate how our method improves performance, we conduct comparison and analysis on effect of entity numbers.
% To analyze the impact of entity numbers, 
Specifically, we first divide the event instances of test set into some subsets based on the number of entities in an event. Since events with a specific number of entities may be too few, results on a subset of a range of entity numbers will yield more robust and convincing conclusion. To make the number of events in all subsets as balanced as possible, we finally get a division of four subsets, whose entity numbers are in the range of [1,3], [4,6], [7,9], and [10,] and event quantities account for 28.4\%, 28.2\%, 25.9\%, and 17.5\%, respectively. 
\begin{figure}[hbt]
    \centering
    \includegraphics[scale=0.29]{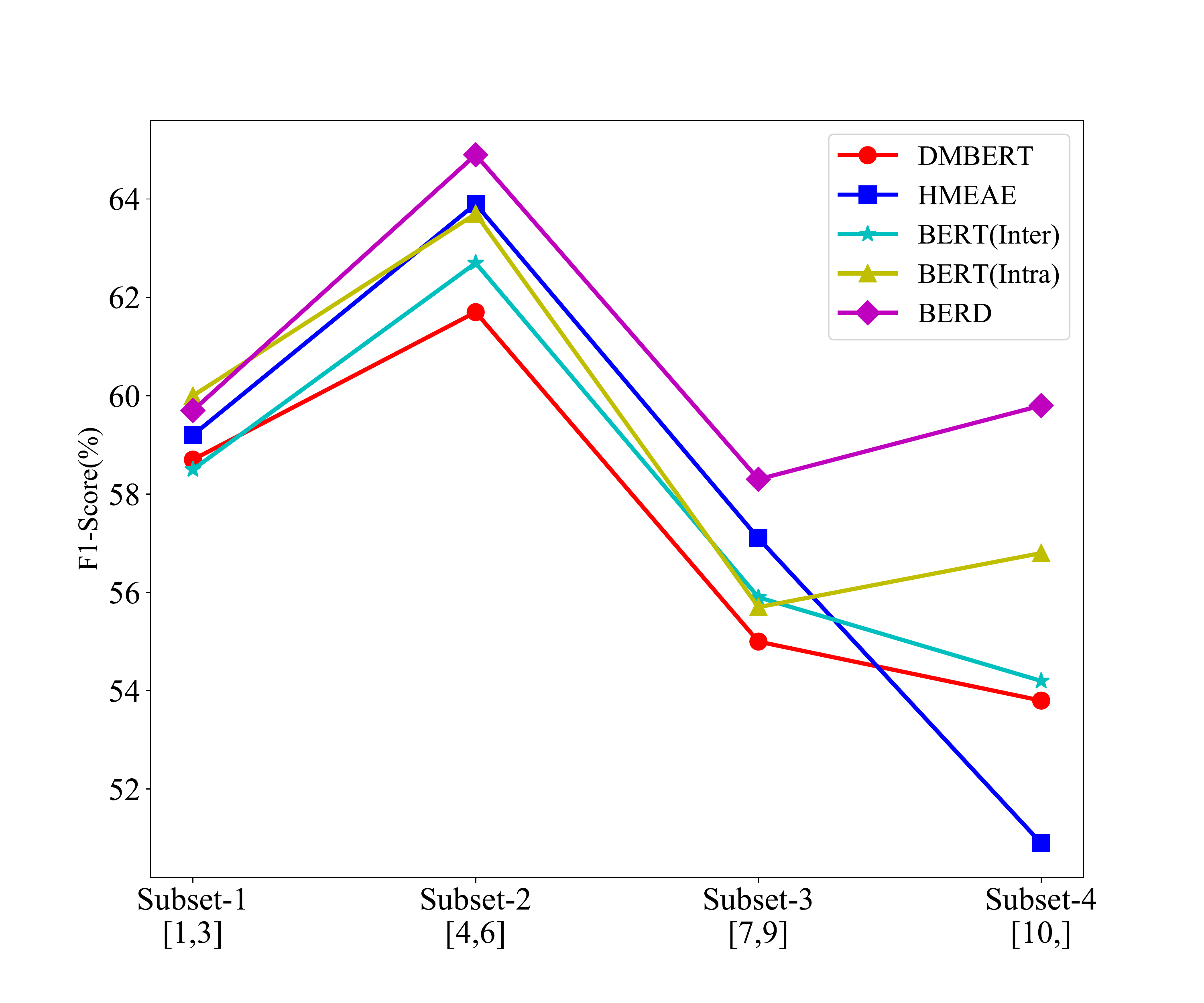}
    \caption{Comparison on four subsets with different range of entity numbers. F1-Score (\%) is listed.
    }
    \label{fig:delta}
\end{figure}

% \begin{table}[hbt]
%     \centering
%     \begin{tabular}{cccc}
%     \hline
%     & \textbf{DMBERT} & HMEAE & \textbf{BERD} \\
%     \hline
%     \emph{Subset-1} &  58.9 & 59.2  &  59.7 \\
%     \emph{Subset-2} & 61.7 & 63.9& 64.9 \\
%     \emph{Subset-3} & 55.0  & 57.1 &  58.4 \\
%     \emph{Subset-4} & 53.8 & 50.9 &59.8  \\
%     \hline
%     \end{tabular}
%     \caption{Performance of DMBERT, BERD and improvements in four subsets. The test set is divided into four subsets with basically the same amount of events.}
%     \label{tab:delta}
% \end{table}

% 结论
% 1. BERD比其他的baseline都更好
The performance of all models on the four subsets is shown in Figure \ref{fig:delta}, from which we can observe a general trend that BERD outperforms other baselines more significantly if more entities appear in an event. More entities usually mean more complex contextual information for a candidate argument, which will lead to a performance degradation. 
BERD alleviates degradation better because of its capability of capturing argument role information of contextual entities.
% 2. BERT(Tensor)也可以缓解这种情况，说明了intra-event argument interaction的有效性
We notice that BERT(Intra) also outperforms DMBERT significantly on \emph{Subset-4}, which demonstrates the effectiveness of intra-event argument interaction.

Note that the performance on \emph{Subset-1} is worse than that on  \emph{Subset-2}, looking like an outlier. The reason lies in that the performance of the first-stage event detection model on \emph{Subset-1} is much poorer (e.g., 32.8 of F1 score for events with one entity).

\subsection{Effect of Overlapping Entity Mentions}
% 引子 
% To explore how the recurrent decoder characters the distribution pattern of multiple argument roles within an event, we analyze the effect of overlapping entities. The test set into two subsets based on whether event mention contains overlapping entity mentions. For each model, the performance on the two subsets is shown in Table \ref{table:ablation_study_nested_entities}, from which we can find that all baselines perform worse on \emph{Subset-Ol} that only contains sentences with overlapping entities. 

Though performance improvement can be easily observed, it is nontrivial to quantitatively verify how BERD captures the distribution pattern of multiple argument roles within an event. In this section, we partly investigate this problem by exploring the effect of overlapping entities. Since there is usually only one entity serving as argument roles in multiple overlapping entities, we believe sophisticated EAE models should identify this pattern.  Therefore, we divide the test set into two subsets (\emph{Subset-O} and \emph{Subset-N} ) based on whether an event contains overlapping entity mentions and check all models' performance on these two subsets.
\begin{table}[hbt]
	\centering
	\begin{tabular}{lcc}
		\hline
		\textbf{Model} & \emph{Subset-O} & \emph{Subset-N} \\
		\hline
		DMBERT & 56.4 & 59.4\\
		HMEAE & 58.8 & 59.6\\
		BERT(Inter) & 57.3 & 58.8\\
		BERT(Intra) & 58.5 & 59.2\\
		BERD & \textbf{60.5} & \textbf{60.1}\\
		\hline
	\end{tabular}
	\caption{Comparison on sentences with and without overlapping entities (\emph{Subset-O} v.s. \emph{Subset-N}). F1-Score (\%) is listed.}
	\label{table:ablation_study_nested_entities}
\end{table}
Table \ref{table:ablation_study_nested_entities} shows the results, from which we can find that all baselines perform worse on \emph{Subset-O}. It is a natural result since multiple overlapping entities usually have similar representations, making the pattern mentioned above challenging to capture.  BERD performs well in both \emph{Subset-O} and \emph{Subset-N}, and the superiority on \emph{Subset-O} over baseline is more significant. We attribute it to BERD's capability of distinguishing argument distribution patterns.

\subsection{Effect of the Bidirectional Decoding}
To further investigate the effectiveness of the bidirectional decoding process, we exclude the backward decoder or forward decoder from BERD and obtain two models with only unidirectional decoder, whose performance is shown in the lines of ``-w/ Forward Decoder'' and ``-w/ Backward Decoder'' in Table \ref{table:ablation_study}.
% Only left- or right-side argument information is utilized for the two scenarios. 
From the results, we can observe that: 
% 2. 双向更好
% (1) Performance of BERD is hindered when forward decoder or backward decoder is excluded (1.6 and 1.3 in terms of F1-Score) due to missing of half-side argument information. Our proposed BERD can simultaneously exploit valuable information of both-side arguments thus gains the most significant improvements. 
(1) When decoding with only forward or backward decoder, the performance decreases by 1.6 and 1.3 in terms of F1 respectively. The results clearly demonstrate the superiority of the bidirectional decoding mechanism
% 1. 都有提高，说明我们的模型本身有效果
(2) Though the two model variants have performance degradation,  they still outperform DMBERT significantly, once again verifying that exploiting contextual argument information, even in only one direction, is beneficial to EAE. 

\begin{table}[hbt]
	
% 	our proposed models on ACE 2005 English corpus. Symbol * indicates the result is adapted from the original paper.
	\centering
	\begin{tabular}{lccc}
		\hline
		\textbf{Model} &  \textbf{P} & \textbf{R} & \textbf{F1} \\
		\hline
		DMBERT & 56.9 & 57.4 & 57.2\\
		\textbf{BERD} & \textbf{59.1} & \textbf{61.5} & \textbf{60.3} \\
	    -w/ Forward Decoder & 58.0 & 59.4 & 58.7\\
		-w/ Backward Decoder & 58.3 & 59.8 & 59.0\\
		-w/ Forward Decoder x2 & 56.8 & 61.1 &  58.9\\
		-w/ Backward Decoder x2 & 57.2 & 61.0 & 59.1\\
		-w/o Recurrent Mechanism & 55.3 & 60.0 & 57.4 \\
% 		\hline
		\hline
	\end{tabular}
	\caption{Ablation study on ACE 2005 dataset (\%).}
	\label{table:ablation_study}
\end{table}
% 3. 为了验证参数，我们进行了两个decoder的实验
% Considering the number of model parameters will be decreased by excluding the forward/backward decoder, we enlarge the two model variants with parameters as same as BERD by introducing another decoder of the same direction respectively (denoted by ``-w/ Forward Decoder x2'' and ``-w/Backward Decoder x2''). Table \ref{table:ablation_study} shows that the enlarged models have similar performance, indicating the improvement come from the complementation of the two decoder with different directions rather than the increment of parameters numbers.

Considering number of model parameters will be decreased by excluding the forward/backward decoder, we build another two model variants with two decoders of the same direction (denoted by ``-w/ Forward Decoder x2'' and ``-w/ Backward Decoder x2''), whose parameter numbers are exactly equal to BERD. Table \ref{table:ablation_study} shows that the two enlarged single-direction models have similar performance with their original versions. 
We can conclude that the improvement comes from complementation of the two decoders with different directions, rather than increment of model parameters.

Besides, we exclude the recurrent mechanism by preventing 
% the previously-predicted argument roles
argument role predictions of contextual entities 
from being fed into the decoding module, obtaining another model variant named ``-w/o Recurrent Mechanism''. The performance degradation clearly shows the value of the recurrent decoding process incorporating argument role information.

\subsection{Case Study and Error Analysis}

\begin{figure}
    \centering
    \includegraphics[scale=0.37]{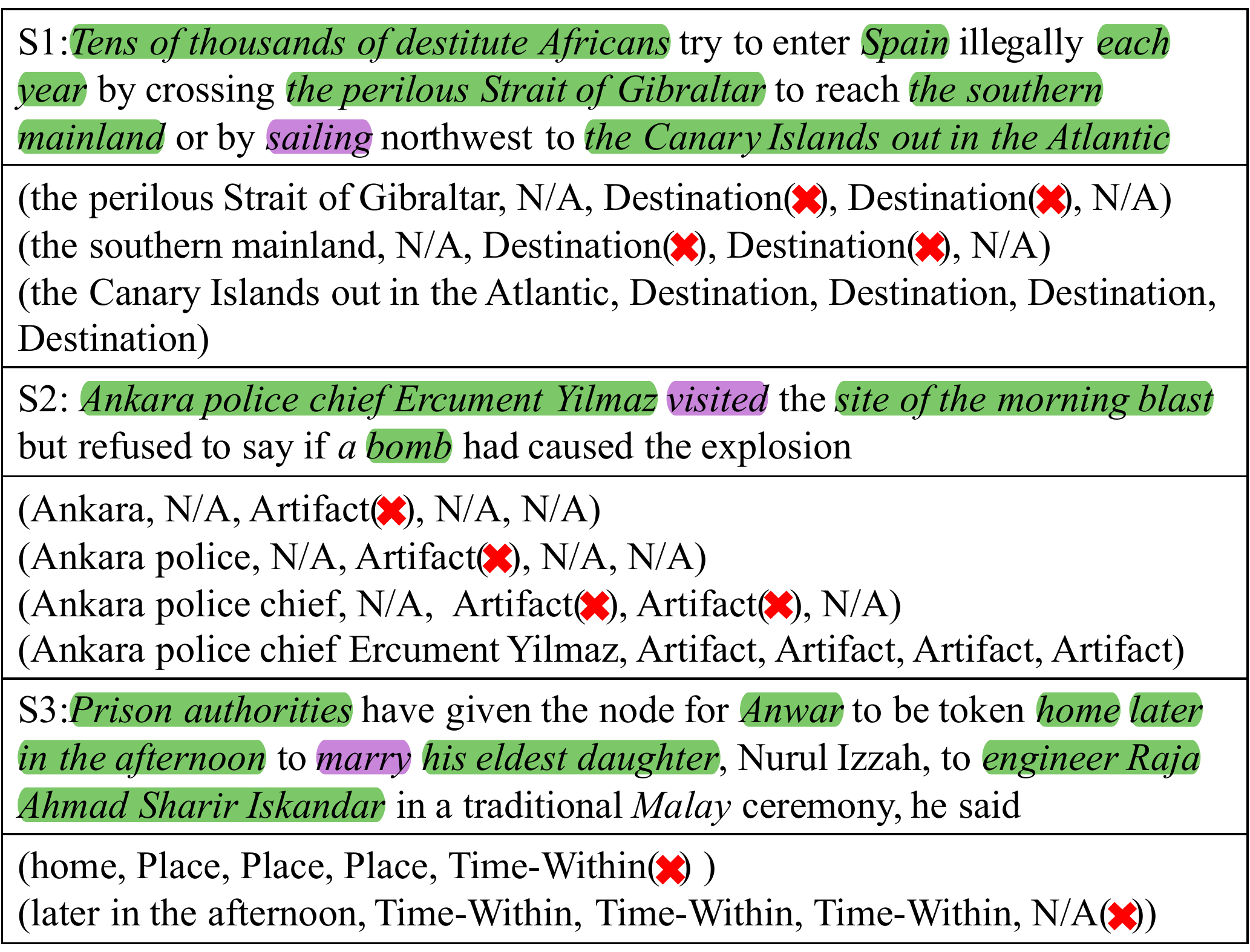}
    \caption{Case study. Entities and triggers are highlighted by green and purple respectively. Each tuple ($E$,$G$,$P_1$,$P_2$,$P_3$) denotes the predictions for an entity $E$ with gold label $G$, where $P_1, P_2$ and $P_3$ denotes prediction of DMBERT, BERT(Intra) and BERD respectively. Incorrect predictions are denoted by a red mark.}
    \label{fig:case_study}
\end{figure}

To promote understanding of our method, we demonstrate three concrete examples in Figure \ref{fig:case_study}. Sentence S1  contains a \emph{Transport} event triggered by ``sailing''. DMBERT and BERT(Intra) assigns \emph{Destination} role to candidate argument ``the perilous Strait of Gibraltar '', ``the southern mainland'' and ``the Canary Islands out in the Atlantic'', the first two of which are mislabeled. It's an unusual pattern that a \emph{Transport} event contains multiple destinations. DMBERT and BERT(Intra) fail to recognize the information of such patterns, 
showing that they can not well capture this type of correlation among prediction results. Our BERD, however, leverages previous predictions to generate argument roles entity by entity in a sentence, successfully avoiding the unusual pattern happening.

% DMBERT assigns ``Time-Within'' role to candidate argumen ``Monday'' and ``Gaza'', the latter of which is mislabeled. It is an unusual pattern that one sentence of an event contains two ``Time-Within'' arguments in the training corpus. DMBERT can not recognize the information of such patterns, since it predicts each entity separately. Our BERD, however, leverages previous predictions to generate argument roles entity by entity in a sentence, successfully avoiding the unusual pattern happening. 

S2 contains a \emph{Transport} event triggered by ``visited'', and 4 nested entities exists in the phrase ``Ankara police chief Ercument Yilmaz''. Since these nested entities share the same sentence context, it is not strange that DMBERT wrongly predicts such entities as the same argument role \emph{Artifact}. Thanks to the bidirectional entity-level recurrent decoder, our method can recognize the distribution pattern of arguments better and hence correctly identifies these nested entities as false instances.
% , which confirms the results and analysis of Table \ref{table:ablation_study_nested_entities}. 
In this case, BERD reduces 3 false-positive predictions compared with DMBERT, confirming the results and analysis of Table \ref{table:ablation_study_nested_entities}. 

% 说误差传播

As a qualitative error analysis, the last example S3 demonstrates that incorporating previous predictions may also lead to error propagation problem. S3 contains a \emph{Marry} event triggered by ``marry''. Entity ``home'' is mislabeled as \emph{Time-Within} role by BERD and this wrong prediction will be used as argument features to identify entity ``later in this after'', whose role is \emph{Time-Within}. As analyzed in the first case, BERD tends to avoid repetitive roles in a sentence, leading this entity incorrectly being predicted as \emph{N/A}. 

\section{Related Work}
We have covered research on EAE in Section \ref{Introduction}, related work that inspires our technical design is mainly introduced in the following.

Though our recurrent decoder is entity-level, our bidirectional decoding mechanism is inspired by some bidirectional decoders in token-level Seq2Seq models, e.g.,  of machine translation \cite{zhou2019synchronous}, speech recognition \cite{chen2020transformer} and scene text recognition \cite{gao2019gate}. 

We formalize the task of EAE as a Seq2Seq-like learning problem instead of a classic classification problem or sequence labeling problem. We have found that there are also some works performing classification or sequence labeling in a Seq2Seq manner in other fields. For example, \citet{yang2018sgm} formulates the multi-label classification task as a sequence generation problem to capture the correlations between labels. \citet{daza2018sequence} explores an encoder-decoder model for semantic role labeling.
% and Alessandro Raganato et al  \cite{raganato2017neural} relies on sequence learning to frame the word sense disambiguation problem. 
We are the first to employ a Seq2Seq-like architecture to solve the EAE task.

\section{Conclusion}

We have presented BERD, a neural architecture with a Bidirectional Entity-level Recurrent Decoder that achieves competitive performance
on the task of event argument extraction (EAE). One main characteristic that distinguishes our techniques from previous works is that we formalize EAE as a Seq2Seq-like learning problem instead of a classic classification or sequence labeling problem. The novel bidirectional decoding mechanism enables our BERD to utilize both the left- and right-side argument predictions effectively to generate a sequence of argument roles that follows overall distribution patterns over a sentence better.

As pioneer research that introduces the Seq2Seq-like architecture into the EAE task, BERD also faces some open questions. For example, since we use gold argument roles as prediction results during training, how to alleviate the exposure bias problem is worth investigating. We are also interested in incorporating our techniques into more sophisticated models that jointly extract triggers and arguments.

\section*{Acknowledgements}
We thank anonymous reviewers for valuable comments. This research was supported by the National Key Research And Development Program of China (No.2019YFB1405802) and the central government guided local science and technology development fund projects (science and technology innovation base projects) No. 206Z0302G.

\bibliographystyle{acl_natbib}
\bibliography{anthology,acl2021}

%\appendix

\end{document}